\documentclass[letterpaper, 10 pt, conference]{ieeeconf}  

\IEEEoverridecommandlockouts                              

\usepackage{times}

\usepackage{soul}
\usepackage{url}
\usepackage[hidelinks]{hyperref}
\usepackage[utf8]{inputenc}
\usepackage[small]{caption}
\usepackage{graphicx}
\usepackage{amsmath}
\usepackage{booktabs}
\usepackage{xcolor}

\usepackage{amsmath,amsfonts,bm}

\def\eqref#1{equation~\ref{#1}}

\def\1{\bm{1}}

\DeclareMathAlphabet{\mathsfit}{\encodingdefault}{\sfdefault}{m}{sl}
\SetMathAlphabet{\mathsfit}{bold}{\encodingdefault}{\sfdefault}{bx}{n}

\usepackage{hyperref}
\usepackage{url}
\usepackage{graphicx}
\usepackage{epsfig}
\usepackage{subcaption}
\usepackage{caption}
\usepackage{wrapfig}
\usepackage{float}
\usepackage{mwe}
\usepackage{xcolor}

\title{\LARGE \bf Investigating Low Data, Confidence Aware Image Prediction on Smooth Repetitive Videos using Gaussian Processes}

\author{Nikhil U. Shinde$^{1}$, Xiao Liang$^{1}$,
Florian Richter$^{1}$, Sylvia Herbert$^{1}$,
and Michael C. Yip$^{1}$ \IEEEmembership{Senior Member, IEEE}
\thanks{$^{1}$ Nikhil U. Shinde, Xiao Liang, Florian Richter, Sylvia Herbert and Michael C. Yip are with the University of California San Diego \{nshinde, x5liang, frichter, sherbert, yip\}@ucsd.edu}}

\begin{document}

\maketitle
\thispagestyle{empty}
\pagestyle{empty}

\begin{abstract}
The ability to predict future states is crucial to informed decision-making while interacting with dynamic environments. 
With cameras providing a prevalent and information-rich sensing modality, the problem of predicting future states from image sequences has garnered a lot of attention. 
Current state-of-the-art methods typically train large parametric models for their predictions. 
Though often able to predict with accuracy these models often fail to provide interpretable confidence metrics around their predictions. 
Additionally these methods are reliant on the availability of large training datasets to converge to useful solutions. 
In this paper, we focus on the problem of predicting future images of an image sequence with interpretable confidence bounds from very little training data. 
To approach this problem, we use non-parametric models to take a probabilistic approach to image prediction. 
We generate probability distributions over sequentially predicted images, and propagate uncertainty through time to generate a confidence metric for our predictions. 
Gaussian Processes are used for their data efficiency and ability to readily incorporate new training data online. 
Our method’s predictions are evaluated on a smooth fluid simulation environment. 
We showcase the capabilities of our approach on real world data by predicting pedestrian flows and weather patterns from satellite imagery.


\end{abstract}

\section{Introduction}

Predicting the future states of an environment is key to enabling smart decision-making. 
As humans, we use predictions to inform daily decisions. 
This ranges from navigating through a crowded crosswalk of pedestrians to using weather forecasts to plan out our week. 
In robotics, we use such predictive models not only to model robots and generate complex control strategies, but also to enable robots to model and better interact with their environment. 
Though perfectly predicting such diverse phenomena requires underlying state information and large trained complex models, humans are often able to make predictions in completely novel environments using imperfect models informed solely by visual input. 
In this paper, we focus on the task of predicting the future in smooth video sequences given only a very limited number of initial frames for context and training. 

With cameras becoming one of the most prevalent sensing modalities, future video prediction is a well-researched topic. 
Several works like \cite{Yu2020Efficient_and_information_preserving_future_farme} and \cite{byeon2018contextvp} directly predict future frames of video in the image space. 
Other works like \cite{finn2016unsupervised} even use these predictions for robotic control. 
However, most of these works use complex parametric models like neural networks that require large datasets to train. 
These works
demonstrate high predictive accuracy, but fail to provide good confidence metrics around their predictions. 
\begin{figure}[t!]
    \centering
    \includegraphics[width=\linewidth]{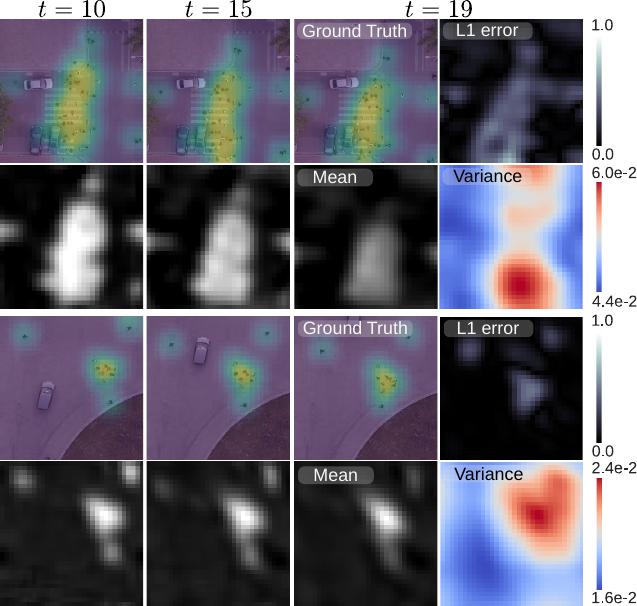}
    \caption{Two real-world pedestrian flow prediction results over time generated by our method trained with only 10 images. 
    Our method predicts future pedestrian motion. 
    Larger variance appears at regions of moving pedestrians, roughly agreeing with regions of error.}
    \label{fig:real_world_pedestrian}
    \vspace{-1.5em}
\end{figure}
In this work we take an initial step towards investigating confidence-aware video prediction from low amounts of data using Gaussian processes (GPs). 
In particular, we look at the problem of predicting the next $t$ frames of a video given the previous $n$ consecutive video frames, given very limited priors and training data. 
Due to the difficulty posed by our restriction to the problem with low training data, which may arise due to constraints such as cost, regulations or entering an unseen scenario, we focus our investigation on predicting smooth videos with highly repetitive, but still complex motion. 
We focus our evaluation on predicting videos of fluid viscosity flows. 
In these videos a large amount of dynamic information can be gained from just a few frames. 
We choose to focus on fluids because the repetitive dynamics and patterns observed in fluids are observed in and used to model many complex real world phenomena such as the flow of pedestrians \cite{helbing1998fluid, following_the_crowd, wang2023fluidbased} and weather prediction \cite{physics_weather_climate_modelling}. 
We showcase our method by demonstrating predictions on real-world examples of crowd flow (Figure~\ref{fig:real_world_pedestrian}) and videos of satellite weather patterns (Figure~\ref{fig:real_world_hurricane}). 
Additionally, due to our very limited training distribution, we expect our models to have limited predictive fidelity. 
To know when we can trust our predictions, we propagate uncertainty through our predictions over time and give confidence metrics on the prediction of future frames.
Our use of GPs for prediction enables high-quality predictions near our easily expandable training distributions, while also providing variance estimates as interpretable metrics around the confidence of our predictions. 

In this paper we present the following novel contributions: 
\begin{enumerate}
	\item A framework for confidence-aware prediction from low data on smooth videos using Gaussian Processes. 
	\item Evaluation of our framework and comparisons on fluid viscosity prediction.
	\item Examples of our framework on real-world data through predictions on the flow of pedestrians and satellite weather patterns. 
\end{enumerate}

\section{Related Works}
Solutions for image sequence prediction problems often heavily rely on large datasets.
There are several available video datasets such as the Kitti \cite{Geiger2013IJRR}, Camvid \cite{camvid_dataset} and Caltech Pedestrian Dataset \cite{caltech_pedestrian_dataset} for prevalent problems like autonomous driving. 
Some video datasets, such as the robot pushing dataset \cite{finn2016unsupervised}, provide video data influenced by external controls for tasks like robot manipulation. 
To solve the problem settings captured by the aforementioned datasets, researchers train large parametric neural networks. 

Most state of the art methods in video prediction build off of a few baseline neural network architectures: convolutional, recurrent and generative models. 
Convolutional neural networks, which rely on learning 2D convolutional kernels, enabled a breakthrough in problems in the image domain \cite{oshea2015introduction_cnns}. 
They have also been extended to problems in video through 3D convolutions \cite{wang2019eidetic_3d_lstm}, \cite{Aigner2018FutureGANAT}. 
Recurrent neural networks (RNNs) \cite{Rumelhart1986LearningRB_RNN_paper} and long short-term memory (LSTM) networks \cite{LSTMS} have more principled architectures to handle the time dependencies that come with sequences of images. 
They have been leveraged by works such as \cite{learning_object_centric_transformation_video_pred}, \cite{wichers2018hierarchical_long_term_video_prediction}, and \cite{walker2017pose_the_pose_knows}. 
Generative adversarial networks directly model the generative distribution of the predicted images \cite{goodfellow2014generative}. 
To handle uncertainty, some researchers have turned to using Variational autoencoders \cite{kingma2014autoencoding} and ensemble networks \cite{ensemble_uncertainty_TALEBIZADEH20114126}. 
Most approaches employ combinations of these architectures to achieve state of the art results. 
Methods like \cite{Yu2020Efficient_and_information_preserving_future_farme} and \cite{byeon2018contextvp} use these methods to directly predict the pixels in future images. 
We take inspiration from this direct prediction approach, along with generative and convolutional approaches, and design our method to directly generate distributions on output pixels while iterating over the image in a kernel-like fashion. 

There is also a large body of work on predicting and simulating fluids. 
The motion of incompressible fluids is governed by the Navier Stokes equations, a set of partially differentiable equations (PDEs). 
Modern techniques such as \cite{RAISSI2019686}, \cite{jiang2020meshfreeflownet}, \cite{greenfeld2019learning_to_optimize_multigrid_pde_solvers} and \cite{li2021fourier_neural_operator} use neural networks to learn to solve complex PDEs from data. 
We compare our method against the FNO2D-time and FNO 3D networks proposed in \cite{li2021fourier_neural_operator}. 

All the methods discussed above focus on a prediction problem where large representative datasets are readily available, while failing to provide confidence metrics around their predictions. 
In this paper, we focus on the case when the available training data is limited to only a few frames and thus understanding predictive confidence even more important. 
In these low-data scenarios, the above methods often fail to converge to useful solutions. 
We choose Gaussian process regression as the core predictive component of our method. 
Gaussian processes have been used across various robotic and machine learning applications to estimate highly nonlinear functions with probabilistic confidence \cite{GP_for_ml_book}. 
These non-parametric models have been regularly used to estimate functions online with very little data. 
\cite{shinde2023objectcentric} use GPs to model complex robot-object interactions online. 
The authors of \cite{GP_for_blimp} use a GP-enhanced reinforcement learning model to learn blimp dynamics online. 
This model improves the state predictions of the traditional ODE modeling approach while also giving a useful uncertainty estimate. 
SOLAR-GP builds upon such system identification approaches and uses localized sparse GP models to learn robot dynamics online to improve teleoperation \cite{SOLAR-GP}. 
PILCO improves the system identification approach further by learning a probabilistic dynamics model \cite{PILCO}. 
They propagate prediction uncertainty through time to facilitate long-term planning and improve policy search methods for reinforcement learning with very little data collection. 
GP’s predictive uncertainty measure has also been widely used by the safety community. 
Safe IML uses GPs to estimate an environment's safety function online \cite{turchetta2019safe_exploration_for_interactive_machine_learning}. 
They leverage the uncertainty outputted by the GP to provide safety guarantees and inform intelligent and risk-aware exploration that does not compromise the robot’s safety. 
In this work, we use GPs for low data, confidence aware predictions of future images from image sequences.

\section{Background: Gaussian Processes}
\begin{figure*}[t]
    \centering
    \includegraphics[width=\textwidth]{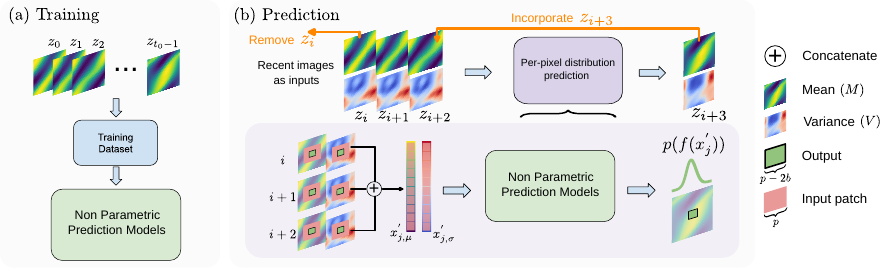}
    \caption{Overview of the proposed method. We begin by pre-processing the initial video frames $[z_{0}, \dots, z_{t_0 - 1}]$ to form the dataset to train a GP regression model. During test time, three sequential input images are processed into test inputs. Our trained model predicts output distributions from these test inputs. These distributions are then combined to form a predictive distribution of the image at the next time step. The prediction is then incorporated into the next set of input images to recursively rollout a sequence of image probabilities.}
    \label{fig:method_overview_graphic}
    \vspace{-1.5em}
\end{figure*}

The core predictive component of our method uses a Single Output GP Regression Model. 
A GP models a function $f$, using training data $(X, f(X))$. 
$X = [x_0, x_1, \dots, x_{n-1}] \in \mathbb{R}^{n \times D}$ are all the training inputs and $f(x)=[f(x_0), \dots, f(x_{n-1})] \in \mathbb{R}^{n \times 1}$ are the training outputs.
Given test inputs $X^{'} \in \mathbb{R}^{m \times D}$ we want to find outputs $f(X^{'})$. 
Let $X_{A} \in \mathbb{R}^{(m+n) \times D}$ refer to all the train and test inputs and $f(X_{A})$ be the corresponding outputs.
A GP relies on the assumption that all the outputs are characterized by a multivariate gaussian distribution $f(X_{A}) \sim \mathcal{N}(\mu(X_{A}), \Sigma_{X_{A}X_{A}})$. 
We assume the mean $\mu(X_{A}) = 0$, and the covariance matrix is characterized by a kernel function $k(x,y)$ such that $\Sigma_{X_{A}, X_{A}}[u,v] = k(X_{A}[u], X_{A}[v])$. 
To solve for the distribution of the test outputs $p(f(X^{'})) \sim \mathcal{N}(\mu(X^{'}), \Sigma_{X^{'}X^{'}})$ we use the marginal likelihood of a multivariate gaussian $p(f(X^{'}| X, f(X), X^{'}))$ to get: 
\begin{equation}\small
    \mu(X^{'}) = k(X^{'}, X)[k(X,X) + \sigma_{n}^2 I]^{-1}f(X)
    \label{eq:GP_mean_pred}
\end{equation}
\begin{equation}\small
    \Sigma_{X^{'}X^{'}} = k(X^{'}, X^{'}) - k(X^{'}, X)[k(X,X) + \sigma_{n}^2 I]^{-1}k(X, X^{'})
    \label{eq:GP_var_pred}
\end{equation}
here, ${\small k(X^{'}, X)[u,v] = k(X^{'}[u], X[v]), \: k(X^{'}, X) \in \mathbb{R}^{n \times n}}$, ${\small k(X^{'}, X) = k(X, X^{'})^T \in \mathbb{R}^{m \times n}}$,  $\sigma_n^2$ is the noise variance, and $I \in \mathbb{R}^{n \times n }$ is the identity matrix. 
To train the GP Regression model we optimize the noise variance, $\sigma^{2}_{n}$, and kernel parameters to maximize the log-likelihood of the training data. 

We use the radial basis function (RBF) kernel: 
\begin{equation}
    k(x,y) = \alpha^2 \exp(-\frac{(x-y)^T \Lambda^{-1}(x-y)}{2})
    \label{eq:rbf_kernel}
\end{equation}
Kernel parameters, $\alpha$ and $\Lambda$, are optimized during training. 
For outputs of dimension $O>1$, we train a GP for each dimension $a \in [0, O-1]$. 
Each model has a kernel $k_a(.,.)$, trained by optimizing its parameters $\sigma_{n,a}, \alpha_{a}$ and $\Lambda_{a}$.

\section{Methods}

\subsection{Problem Statement and Prediction Framework}
We define an image sequence as a sequence of frames $[z_{0}, z_{1} \dots z_{t_{0} - 1}, z_{t_{0}}, \dots z_{t-1}, z_{t} \dots]$. 
Here $z_{i} \in \mathbb{R}^{H \times W}$ denotes the $i$-th frame in the sequence. 
Given initial training frames $[z_{0}, \dots z_{t_{0} - 1}]$, our objective is to predict frames  $[z_{t_{0}}, \dots z_{t}, \dots]$.
As additional frames $[z_{t_{0}}, \dots z_{t^{'}}]$ become available, they may be incorporated into the model’s training data to improve the accuracy of the future prediction. 

Figure \ref{fig:method_overview_graphic} provides an overview of our method. We train a model to use recurring motion patterns to understand scene dynamics to predict future images. 
Our model learns and predicts on square image patches of dimension $(p, p)$. 
Predicting at a smaller scale enables us to better use our limited training data and extract smaller repeating patterns that are more likely to recur across space and time. 
Our method predicts one frame at a time given the $3$ most recent, seen, and predicted frames. 
We use $3$ input frames to capture second-order dynamics. 
The GP regression models predict per-pixel distributions in future images. 
These are combined to form a random variable image. 
The predicted image is incorporated into the next set of inputs, and the process is repeated. 
Our method must handle a combination of random and known inputs, while propagating probability distributions through time.

\subsection{Training} \label{section_training}
To construct our model, we first create a training data set from frames $[z_{0}, \dots, z_{t_{0}-1}]$.
We divide the images into sets of $4$ sequential images $[z_{i}, z_{i+1}, z_{i+2}, z_{i+3}], i \in [0, t_{0} - 4]$. 
To create a datapoint we take $p$ dimensional patches corresponding to the same pixel locations from each image. 
$z_{i}[k:k+p, l:l+p] \in \mathbb{R}^{p \times p}$ denotes a $p$ by $p$ patch in image $z_{i}$ starting at pixel $(k,l)$. 
A training input, $x_{j} \in \mathbb{R}^{3p^{2}}$, is created by flattening and concatenating the patches from the first $3$ images. 
The corresponding training output, $f(x_{j}) \in \mathbb{R}^{(p-2b)^2}$, is created by flattening the corresponding patch from the $4$th image $z_{i+3}$, cropped with a patch boundary term $b$: $z_{i+3}[k+b:k+p-b, l+b:l+p-b] \in \mathbb{R}^{(p-2b) \times (p-2b)}$. 
When $b>0$, we do not predict the outer edges of the patch, where predictions may suffer due to contributions from the scene outside our input.
Within each set of $4$ sequential images, we select data points with a stride of $s$ pixels. 
In this paper, we use a ``wrapping" approach to handle patches that extend beyond the edge of an image. 
This approach assumes the frame $z_{i}$  captures a periodic environment where $z[H + i,W + j] = z[i-1, j-1]$ and $z[-i, -j] = z[H - (i+1), W - (j+1)]$.
Approaches like padding frames or skipping incomplete patches are possible, but not further explored in this paper. 
We repeat this procedure for every set of images to create the training dataset with $n$ data points: $(X, f(X)) = (x_{j},f(x_{j})), j \in [0, n-1]$. $X \in \mathbb{R}^{n \times 3p^2}$ and $f(X) \in \mathbb{R}^{n \times (p-2b)^2}$. 

We create a GP Regression model for every output dimension $O = (p-2b)^2$. 
Each model is trained by optimizing its noise, $\sigma_{n, a}$, and kernel parameters, $\alpha_{a}$ and $\Lambda_{a}$ in $k_{a}(.,.)$, to maximize the log likelihood of the training data for output dimension $a \in [0, (p-2b)^2]$. 
To predict future images, each GP model outputs a mean and variance corresponding to a pixel in the output patch.
The predicted image $z_{i}$, is represented by a mean and variance image pair $(M_{i}, V_{i})$. 
Each pixel in $M_i$ and $V_i$ corresponds to the mean and variance of the predicted random gaussian variable for that pixel location, respectively. 

\subsection{Prediction} \label{section_prediction}
Once trained, we can use our model to rollout predictions for any $T$ time steps into the future, starting from $3$ known, consecutive input images $[z_{i}, z_{i+1}, z_{i+2}]$. 
We use a recursive method to predict multiple frames into the future. 
Our model takes the $3$ most recently seen and predicted frames and uses them to predict one future frame, represented as a random variable. 
We incorporate this predicted random variable as the latest image in the $3$ frame inputs to predict the next time step. 
This process is repeated to predict the desired $T$ steps into the future. 
We begin by discussing our predictions in the context of predicting the fourth time step and onwards. 
Starting at the fourth prediction, all the input images are random variables previously outputted by our model.
The first three predictions incorporate known, observed input images and predicted random input images. 
These initializing predictions will be discussed as a special case of the more general prediction from all random variable input images.  

We discuss the general method of predicting $z_{i+3}$ from input images $[z_{i}, z_{i+1}, z_{i+2}]$, which are all random variables outputted by our model. 
To predict $z_{i+3}$, we create a set of $m$ test inputs ${\small x^{'}_{j}, \, j \in [0, m-1]}$.  
Each test input is a multivariate Gaussian random variable composed of $3p^2$ independent Gaussian random variables selected from the input images. 
Since the input images are predicted random variables, our test inputs are selected from their corresponding mean and variance images: ${\small [(M_{i}, V_{i}), (M_{i+1}, V_{i+1}), (M_{i+2}, V_{i+2})]}$.
We make simplifying assumptions that the predictions of each GP model as well as the outputs of the same model on different inputs are independent. 
Without assuming independence, we would have the computationally intractable task of tracking the covariance across all pixels across all time. 
As a result, the predicted images and their sub-patches can be flattened, concatenated, and represented as one multivariate Gaussian random variable.  
We use the patch-based selection method described in Section \ref{section_training}, separately, on the sets of consecutive mean and variance images to generate the mean and variance input vectors, $x^{'}_{j, \mu} \in \mathbb{R}^{3p^2}$ and $x^{'}_{j, \sigma} \in \mathbb{R}^{3p^2}$ respectively. 
These vectors specify the multivariate gaussian distribution of the input ${\small x^{'}_{j} \sim \mathcal{N}(x^{'}_{j, \mu}, \Sigma_{x^{'}_{j, \sigma}})}$. 
To construct $\Sigma_{x^{'}_{j, \sigma}}$, we use our independence assumptions such that the input covariance is a diagonal matrix with the vector of variances, $x^{'}_{j, \sigma}$, along the diagonal. 
We adjust our stride to generate an input to predict every pixel in the future image. 

\begin{figure*}[t]
    \vspace{1.5mm}
    \centering
    \includegraphics[width=\textwidth]{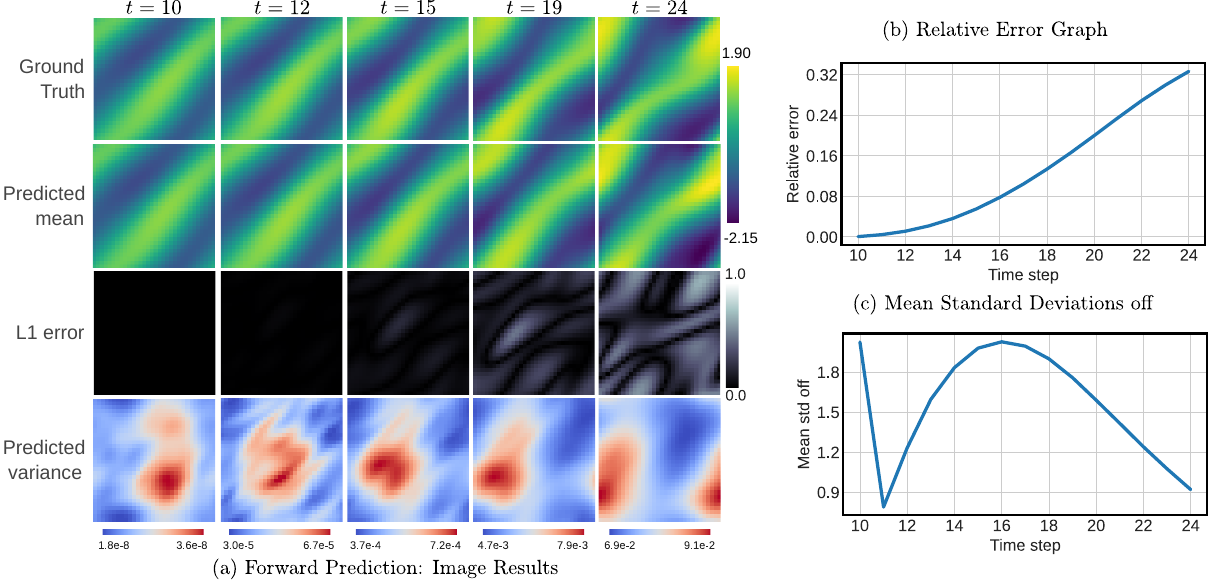}
    \caption{Forward Prediction Experiment:
    Our model, trained using frames $[z_{0}, \dots, z_{9}]$, is used to predict frames $[z_{10}, \dots, z_{24}]$ of a 2D Navier Stokes simulation. 
    (a) Ground truth, predicted mean, l1-error, and variance images.
    (b-c) Graphs of the prediction's relative error (RE) and mean standard deviations off (StdE). 
    This shows our model's ability to accurately predict dynamic scenes. Error and variance increases overtime. 
    Before $t=19$, the variance effectively informs the erroneous region. 
    At later timesteps the variation in the variance becomes less informative.
    The base variance increases with the model becoming more uncertain overall. 
    The accuracy of the model’s confidence in its own predictions naturally oscillates as seen in the mean std off plot in Figure.~\ref{fig:simple_prediction_experiment}c. 
    The decrease in this metric is caused when the model’s predicted variance grows faster than the true-error, while an increase correlates to the inverse.
    The increase in the predicted variance eventually dominates true error causing mean stds off to decrease, which is preferable as it ensures we provide a conservative estimation where the predicted bounds capture the ground truth. 
    }
    \label{fig:simple_prediction_experiment}
    \vspace{-1.5em}
\end{figure*}

We discuss our method in the context of predicting a single output dimension $a \in [0, (p-2b)^2]$ from a single input $x^{'}_{j}$. 
Our model predicts the random variable $f(x^{'}_{j})[a]$. 
As in standard GP Regression, we are solving for the distribution of $p(f(x^{'}_{j})[a])$. 
We solve for $p(f(x^{'}_{j})[a])$ by marginalizing $p(f(x^{'}_{j})[a]| X, f(X), x^{'}_{j})$ over the input image distributions.
\begin{equation}
    \begin{gathered}
        p(f(x^{'}_{j})[a]) = \int^{\infty}_{-\infty} p(f(x^{'}_{j})[a]| x^{'}_{j}, X, f(X)[:, a]) p(x^{'}_{j}) dx^{'}_{j}
    \end{gathered}
    \label{eq:intractable_integral}
\end{equation}
Solving this integral is analytically intractable.
We approximate the posterior distribution from \eqref{eq:intractable_integral} to be Gaussian. 
Having the outputs form a multivariate Gaussian, like the inputs, enables recursive prediction. 
To solve for $p(f(x^{'}_{j})[a] )$ we take advantage of this assumption and use moment matching in a method akin to \cite{PILCO}.
Our method is distinguished from \cite{PILCO} in its use of multiple past states as inputs, prediction on images, and incorporation of known input states. 
Moment matching enables us to directly solve for the mean $\mu(x^{'}_{j})[a]$ and variance $\Sigma(x^{'}_{j})[a,a]$ of the outputted Gaussian distribution. 
This gives us the following formula to predict the mean of an output pixel from all random inputs:
\begin{equation}
    \begin{gathered}
        \mu(x^{'}_{j})[a] = d_a^T \beta_a  \\
        d_a[i] = \alpha_a^2  (|\Sigma_{x^{'}_{j, \sigma}} \Lambda_a^{-1} + I|)^{-\frac{1}{2}}  e^{-\frac{1}{2} v_{i}^{T}(\Sigma_{x^{'}_{j, \sigma}} + \Lambda_a)^{-1}v_{i}} \\
        \beta_a = [k_a(X, X) + \sigma^{2}_{n} I]^{-1} f(X)[:, a]
    \end{gathered}
    \label{eq:mean_propagation_random}
\end{equation}
Here $d_{a} \in \mathbb{R}^{n}$.
$v_{i} = x^{'}_{j, \mu} - x_{i}$ where $x_{i}$ is the $i^{\text{th}}$ training input. 
To predict variance from random inputs we use: 
\begin{equation}
    \begin{gathered}
        \Sigma(x^{'}_{j})[a,a] = \alpha_{a}^2 - \text{Tr}\left( (k_a(X, X) + \sigma_{n, a}^{2} I)^{-1} Q_{aa}\right) + \\
        \beta_a^T Q_{aa} \beta_a - \mu(x^{'}_{j})[a]
    \end{gathered}
    \label{eq:var_propagation_random}
\end{equation}
\begin{equation}\small
    \begin{gathered}
        Q_{aa}[i,k] = k_a(x_{i}, x^{'}_{j, \mu}) k_a(x_{k}, x^{'}_{j, \mu}) |R|^{-\frac{1}{2}} e^{\frac{1}{2}z_{ik}^T R^{-1}\Sigma_{x^{'}_{j, \sigma}} z_{ik}}
    \end{gathered}
    \label{eq:var_propagation_random_Qmatrix}
\end{equation}
Here $Q_{aa} \in \mathbb{R}^{n \times n}$, $R =  \Sigma_{x^{'}_{j, \sigma}}2\Lambda_a^{-1} + I \in \mathbb{R}^{n \times n }$ and $z_{ik} = \Lambda_a^{-1} v_i + \Lambda_a^{-1}v_k$. 
These equations are used on all the test inputs to predict the mean and variance for every pixel in $z_{i+3}$. 
The predicted image $z_{i+3}$ is stored as a mean and variance image tuple: $(M_{i+3}, V_{i+3})$. 
To continue the predictive rollout, we incorporate the latest prediction into a new set of input images $[z_{i+1}, z_{i+2}, z_{i+3}]$ to predict $z_{i+4}$. 
The mean images $[M_{i}, \dots, M_{i+3}, \dots]$ act as our predictions, while the variance images  $[V_{i}, \dots, V_{i+3}, \dots]$ act as a confidence measure on our prediction. 


In the first $3$ predictions some or all of the input images are known, observed images. 
Predictions with these inputs are special cases of the general formulation with all random variable inputs. 
To solve this case, we still consider all components of our input as random variables. 
We use our independence assumptions to disentangle the predictive method into parts that solely interact with the input dimensions contributed from the known observed images, $x^{'K}_{j}$, and those contributed from the predicted random variable images, $x^{'R}_{j}$. 
We treat the predicted random variable component of the input as a multivariate Gaussian, $x^{'R}_{j} \sim \mathcal{N}(x^{'R}_{j, \mu}, \Sigma_{x^{'R}_{j, \sigma}})$, and use a Dirac delta, $\delta(x - x^{'K}_{j, \mu})$, to describe the distribution at the observed components. 
We split up the RBF kernel function $k_{a}(x,y) = k^{K}_{a}(x^{K}, y^{K}) k^{R}_{a}(x^{R}, y^{R})$ into components that interact solely with the known, $k_{a}^{K}$, and random, $k_{a}^{R}$ dimensions of the inputs. 
These known and random kernels have parameters $\alpha_{a,K}, \Lambda_{a,K}$ and $\alpha_{a,R}, \Lambda_{a,R}$ respectively. 

We use moment matching with these representations to solve the intractable integral from \eqref{eq:intractable_integral} for distributions on future predictions. 
The delta distribution at the known pixels allows us to integrate out certain contributions from the known components during the moment matching steps. 
This gives us the following equation to predict the mean from these hybrid inputs: 
\begin{equation} \label{eq:mean_propagation_hybrid}
    \begin{gathered}
        \mu(x^{'}_{j}) = d_{a, h}^{T} \beta_a \\
        d_{a, h}[i] = k^{K}_{a}(x^{K}_{i}, x^{'K}_{j,\mu}) \cdot \\
        \frac{\alpha_{a,R}^{2}}{\sqrt{|\Sigma_{x^{'R}_{j, \sigma}} \Lambda_{a, R}^{-1} + I|}} 
        e^{-\frac{1}{2} v_{i,R}^{\top} \left( \Sigma_{x^{'R}_{j, \sigma}} + \Lambda_{a, R} \right)^{-1} v_{i,R}}
    \end{gathered}
\end{equation} 
Where the $\cdot$ operator represents elementwise multiplication. 
$\beta_{a}$ is defined in \eqref{eq:mean_propagation_random}, and 
$v_{i,R} = x^{'R}_{j, \mu} - x^{R}_{i}$. 
Here $x^{K}_{i}, x^{R}_{i}$ are the components of the $i^{\text{th}}$ training data input that correspond to the known and random components of the test input respectively. 
To predict the variance from the hybrid inputs we use: 
\begin{equation}\label{eq:var_propagation_hybrid}\small
    \begin{gathered}
        \Sigma(x^{'}_{j})[a,a] = \alpha_{a}^2 - \text{Tr}((k_{a}(X, X) + \sigma_{n, a}^2 I)^{-1} Q_{aa, h}) + \\
        \beta_{a}^{T} Q_{aa, h} \beta_{a} - \mu(x^{'}_{j})[a] \\
        Q_{aa, h} = Q_{aa, K} \cdot Q_{aa, R} \\
        Q_{aa, K} = k^{K}_{a}(x^{'K}_{j, \mu}, X^{K})^{T} 
        k_{a, K}(x^{'K}_{j, \mu}, X^{K}) \in \mathbb{R}^{n \times n} \\
        Q_{aa, R}[i, k] = \frac{1}{\sqrt{|R_{R}|}}  \\
        k^{K}_{a}(x^{R}_{i}, x^{'R}_{j, \mu}) k^{R}_{a}(x^{R}_{k}, x^{'R}_{j, \mu})  e^{\frac{1}{2} z_{ik, R}^{T} R_{R}^{-1} \Sigma_{x^{'R}_{j, \sigma}} z_{ik, R}}
    \end{gathered}
\end{equation}
The $\cdot$ operator represents elementwise multiplication. 
$X^{K}$ are the dimensions associated with the known inputs across all training inputs. 
$R_{R} = \Sigma_{x^{'R}_{j, \sigma,}}(2\Lambda_{a,R}^{-1}) + I$, and $z_{ik, R} = \Lambda_{a,R}^{-1}v_{i,R} + \Lambda_{a, w}^{-1}v_{k,R}$.
Additionally, $\alpha_{a}, \sigma_{n,a}$ are parameters of the original (unsplit) kernel function $k_{a}$. 
Together we use \eqref{eq:mean_propagation_random}, \eqref{eq:var_propagation_random}, \eqref{eq:mean_propagation_hybrid} and \eqref{eq:var_propagation_hybrid} to predict the mean and confidence bounds on future images.

\section{Experiments and Results}\label{section_experiments_and_results}

We test our methods by predicting the vorticity of an incompressible fluid in a unit torus environment. 
Our data is computed using the 2D Navier Stokes Equations. 
We generate our data with traditional PDE solvers using the code and approach detailed in \cite{li2021fourier_neural_operator}. 
The fluid simulation generates image sequences whose pixels change smoothly over both space and time. 
This environment is well suited to the RBF kernel. 
The dynamics of the toroidal environment wrap around the square image frame. 
This enables us to utilize the “wrapping” approach when creating edge patches. 
Each image pixel is a float centered at 0. 
We directly predict future pixels using our 0-mean GP regression model. 

As shown in Figure~\ref{fig:simple_prediction_experiment}(a),  each image $z_{t} \in \mathbb{R}^{H \times W}$ in the image sequence represents the vorticity of the fluid at time $t$. 
We fix our fluid viscosity at $1e-3$, time step at $1e-4$ seconds and image resolution at $H=W=32$. 
Our experiments use input patch dimension $p=15$ with patch boundary $b=7$, resulting in $(1,1)$ output patches. 
We use a single GP Regression model to predict this single output dimension. 
We generate training inputs using a stride of $s=2$ and test inputs using a test stride of $s=1$. 
Each experiment predicts $15$ frames into the future. 
The training images and initial input images are specified for each experiment.

We use two metrics: relative error \cite{li2021fourier_neural_operator} ($RE(z, \tilde{z})$) and mean standard deviations off ($StdE(z,\tilde{z_{\mu}}, \tilde{z_{\sigma}})$) to analyze the performance of our model. The relative error is given by
\begin{equation}
    \begin{gathered}
        RE(z, \tilde{z}) = \frac{||z - \tilde{z}||_{2}}{||z||_{2}}
    \end{gathered}
    \label{eq:relative_error_eqn}
\end{equation}
where $z \in \mathbb{R}^{H \times W}$ is the ground truth image, $\tilde{z} \in \mathbb{R}^{H \times W}$ is the predicted image, and $||.||_{2}$ is the 2-norm. 
This normalizes the error with the magnitude of the original image.

Mean standard deviations off is given by 

\begin{equation}
    \begin{gathered}
        StdE(z, \tilde{z_{\mu}}, \tilde{z_{\sigma}}) = \frac{1}{H \cdot W}\sum_{i=0}^{H-1} \sum_{j=0}^{W-1} \left( \frac{|z[i,j]-\tilde{z_{\mu}}[i,j]| }{\sqrt{\tilde{z_{\sigma}}[i,j]}} \right),
    \end{gathered}
    \label{eq:avg_stdsoff_eqn}
\end{equation}
where $z \in \mathbb{R}^{H \times W}$ is the ground truth image,   $\tilde{z_{\mu}}, \tilde{z_{\sigma}} \in \mathbb{R}^{H \times W}$ is the predicted mean and variance images, and $|.|$ is the absolute value function. 
This metric provides the average absolute standard deviations between our predicted mean image and the ground truth. 

\textbf{Forward Prediction Experiment:} \label{section_forward_pred_experiment}
We train our model using the first $t_{0}=10$ frames of a video $[z_{0}, \dots z_{9}]$. 
This model is used to predict the next $15$ frames $[z_{10}, \dots, z_{24}]$, from input images $[z_{7}, z_{8}, z_{9}]$. 
The results of this experiment are shown in Figure \ref{fig:simple_prediction_experiment}. 
Our model's predicted mean images track the complex dynamics of the ground truth with very little training data. Figure~\ref{fig:simple_prediction_experiment}a indicates that our model predictions are relatively accurate. 
Despite the error increases over time shown by the error images and Figure~\ref{fig:simple_prediction_experiment}b, our method's uncertainty also increases as shown by the variance images. 
Note that our model's predicted variance is more trustworthy at earlier prediction time points (e.g. $t \le 19 $), agreeing with the error image. 
However, relative spatial variations become ineffective when predicting far into the future.
A graph of $StdE$ in Figure~\ref{fig:simple_prediction_experiment}c shows the oscillation in the accuracy of the model's confidence in its predictions. 
$StdE$ decreases when the model's predicted variance grows faster than the true error. 
Meanwhile a lower predicted variance but larger true error results in this metric increasing. 
In Figure~\ref{fig:simple_prediction_experiment}c we can see that the growth of the predicted variance eventually dominates the true error causing the $StdE$ to decrease. 
This behavior is preferable as a lower $StdE$ ensures our predictions provide a conservative estimate and our predicted bounds capture the ground truth.

\begin{figure}[t!]
    \vspace{1.2mm}
    \raggedright
    \begin{subfigure}[b]{0.47\textwidth}
        \raggedright
        \includegraphics[width=\textwidth]{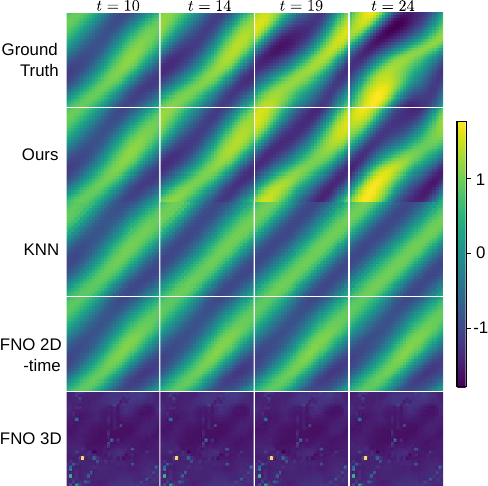}
        \caption{Predictive Comparison Images}
        \vspace{2mm}
        \label{fig:predictive_comparison_images}
    \end{subfigure}
    \hfill
    \begin{subfigure}[b]{0.45\textwidth}
        \raggedright
        \includegraphics[width=0.96\textwidth]{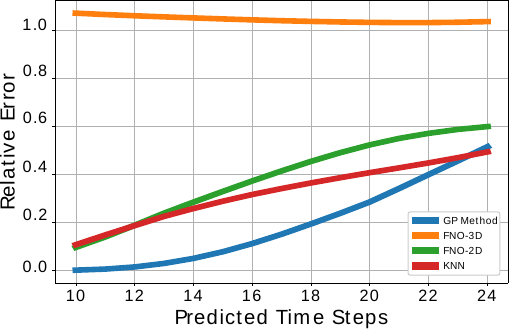}

        \caption{Predictive Comparison Errors}
        \label{fig:predictive_comparison_errors}
    \end{subfigure}
    \caption{Predictive Comparison Experiment:
    This figure compares the predictions on 2D Navier Stokes simulations between our method, a non-parametric KNN-based method, and the neural network-based methods, FNO-2D-time and FNO-3D trained on similarly low data. (a) Snapshots from predictions on a single test sequence.
    (b) Relative error vs the predictive time step averaged across 100 prediction tests.
    }
    \vspace{-1em}\label{fig:predictive_comparison_experiment}
\end{figure}

\begin{figure*}[t]
    \vspace{1.2mm}
    \centering
    \includegraphics[width=0.97\linewidth]{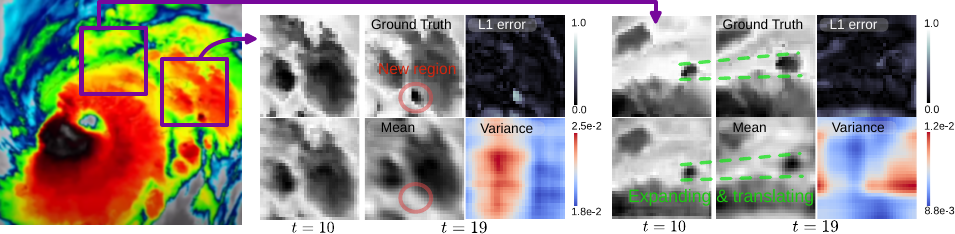}
    \caption{Real-world hurricane satellite image results generated by our method trained with only 10 images. Training and prediction are performed on gray-scale satellite image patches. A snapshot of the satellite video and selected patches are shown on the left, whereas prediction results are shown in the middle and right. Our method can predict translation dynamics and relatively complex dynamics such as a region expanding. Note that it cannot predict trends that are not present in training, for example, emergence of new regions.}
    \vspace{-1.5em}
    \label{fig:real_world_hurricane}
\end{figure*}

\textbf{Predictive Comparison Experiment:}
We compare our method’s performance to the FNO2D-time and FNO 3D neural network methods in \cite{li2021fourier_neural_operator}. 
All methods are trained using a similarly low number of training images. 
We also contrast with a non-parametric K Nearest Neighbors approach. 
We set $k=20$, use the same input and output pre-processing, and use the same optimized RBF kernel-based similarity metric to provide a fair comparison. 
The results are in Figure~\ref{fig:predictive_comparison_experiment}a. 
Both the neural networks and the KNN fail to learn the complex dynamics from the few available training frames, and simply output noise or sequences of identical frames. 
Figure~\ref{fig:predictive_comparison_experiment}b shows the relative error averaged over prediction experiments on 100 different Navier Stokes simulations. 


\textbf{Real World Experiments}: We evaluate the proposed method on the image prediction task of pedestrians on a public dataset \cite{pedestrian_dataset}. 
The dataset contains videos of pedestrians and vehicles taken from an overhead camera. 
We create grey-scale images representing pedestrians' positions by dilating and smoothing their pixel coordinates on images. Figure~\ref{fig:real_world_pedestrian} shows the prediction results on the grey-scale images. 
The proposed approach can predict pedestrian motion trends. At the end of the prediction, variance is higher in regions of larger pedestrian motion.
Higher-variance regions overlap with regions of larger error.

In addition, our method is evaluated with a real satellite video of the hurricane Ian \cite{Denver7}. An example hurricane image and related results are shown in Figure~\ref{fig:real_world_hurricane}. 
We crop out patches of the video and convert them to grey-scale. 
Our method can capture interesting dynamics such as translation and expansion of features. 
As expected, it fails at predicting trends that are not presented in the training images, such as the emergence of new features. 
The predicted variance is larger at regions that have high-intensity variation, indicating that our method becomes more uncertain when the dynamics are complex.


\section{Conclusion and Future Discussion}

In this paper we provided a novel method using non-parametric GP-Regression models for confidence aware prediction of future images in an image sequence with very little training data. 
We evaluated our method on predictions of a 2D Navier Stokes simulation environment. 
These experiments demonstrated our method's ability to confidently capture the ground truth image sequence within our predicted image distribution for complex dynamic environments. 
We showcased our approach on real world environments by predicting pedestrian traffic flows as well as satellite weather phenomenon. 
This demonstrates our method's ability to be applied to real world applications, especially tasks where collecting a large representative dataset may be difficult due to constraints from cost or regulation restrictions. 


This work is an initial step into using GPs to predict images with interpretable confidence metrics. 
More research is needed on using GPs for image prediction to improve their abilities to learn these complex visual dynamics, enhance their predictive accuracy and tighten their predicted confidence bounds. 
We also seek to explore avenues to combine the benefits of our approach, such as the ability to cater towards newly acquired small batches of data and provide interpretable confidence metrics with the accuracy and sharpness benefits that come from big data and large parametric model driven approaches.
Finally we hope to use the predictions from these approaches and their confidence metrics to better guide online decision making for autonomous robotic systems.

\bibliographystyle{IEEEtran}
\bibliography{ijcai22}

\clearpage

\appendix
\section{Appendix}
\subsection{Predictions: The first $3$ Predictions in a sequence of predictions} \label{section_hybrid_input_prediction}

The first three predictions of a rollout involve predicting from a combination of known and predicted random variable sets of input images. 
We treat these predictions as special cases of the more general case presented in Section \ref{section_prediction}. 
Each input $x^{'}_{j}$ is composed of $3p^2$ independent random variables, with $p^2$ random variables contributed from each input image. 
We separate the input $x^{'}_{j}$ along the dimensions of the input that are known and random, to handle the different components separately. 
$x^{'}_{j, Rn}$ and $x^{'}_{j, Kn}$ are random variables that denote the random and known components of the input respectively. 
$X_{Rn}, X_{Kn}$ and $x_{i, Rn}, x_{i, Kn}$ reference the corresponding known and random dimensions in all the training inputs and a single training input respectively. 

The kernel functions $k_{a}$ and the probability distribution over the input $p(x^{'}_{j})$ are the only forms of interaction with the inputs while predicting the output distribution $p(f(x^{'}_{j})[a])$. 
The structure of these functions and our independence assumptions allow us to cleanly split up our inputs into known and random components. 
We use the Radial Basis Function (equation \ref{eq:rbf_kernel}) as our kernel function $k_{a}(x,y)$. 
In this function the inputs interact with one another along the same dimension, allowing us to re-write the kernel as: 
\begin{equation}
    \begin{gathered}
        k_a(x,y) = \alpha_a^2 \exp \left( -\frac{(x-y)^T \Lambda_a^{-1}(x-y)}{2}) \right) 
        = \alpha_{a,Kn}^2 \cdot \\
        \exp\left(-\frac{(x_{Kn}-y_{Kn})^T \Lambda_{a, Kn}^{-1}(x_{Kn}-y_{Kn})}{2}\right) \cdot \\
        \alpha_{a,Rn}^2 \exp \left( -\frac{(x_{Rn}-y_{Rn})^T \Lambda_{a, Rn}^{-1}(x_{Rn}-y_{Rn})}{2} \right) \\
        = k_{a, Kn}(x_{Kn}, y_{Kn}) \cdot k_{a, Rn}(x_{Rn}, y_{Rn})
    \end{gathered}
    \label{eq:split_rbf_kernel}
\end{equation}
The subscripts on the inputs correspond to their known and random dimensions. 
$k_{a, Rn}$ and $k_{a, Kn}$ are the kernel functions that act on the known and random dimensions respectively. 
Each are parameterized by their own set of kernel parameters: $\alpha_{a, Kn}, \Lambda_{a, Kn}$ and $\alpha_{a, Rn}, \Lambda_{a, Rn}$ respectively. 
$\alpha_{a, Rn} \cdot \alpha_{a, Kn} = \alpha_{a}$. 
$\Lambda_{a, Kn}$ and $\Lambda_{a, Rn}$ are block diagonal matrices sampled from $\Lambda_{a}$, a large diagonal matrix, along the known and random dimensions. 
Using our assumption that all predicted pixels within and across images are independent, we separate the probability distribution: 
\begin{equation}
    \begin{gathered}
        p(x^{'}_{j}) = p(x^{'}_{j, Kn}) \cdot p(x^{'}_{Rn})
    \end{gathered}
    \label{eq:split_probability_distribution}
\end{equation}
$p(x^{'}_{j, Rn})$ is a multivariate gaussian distribution of the random components of the input and $p(x^{'}_{Kn})$ denotes the joint distribution of all the known input pixels. 

We construct our inputs according to this split input structure. 
The random component of each input is a multivariate gaussian distribution,
$x^{'}_{j, Rn} \sim \mathcal{N}(x^{'}_{j, \mu}, \Sigma_{x^{'}_{j, \sigma}})$, 
specified by its mean and covariance matrix. 
To create $x^{'}_{j, \mu}$ and $\Sigma_{x^{'}_{j, \sigma}}$ for the random test inputs, we use the method described in Section \ref{section_prediction}.
We sample and concatenate patches from the mean and variance images corresponding to the subset of input images that are random variables.  
Pixels in the analogous patches of the known input images are flattened and concatenated to form $x^{'}_{j, Kn, \mu}$, the observed values of $x^{'}_{j, Kn}$ the random variable that is the known component of the test input. 
If all the input images are known, the test inputs are created in the same manner as the training inputs, as described in section \ref{section_training}

Our model output is the predicted random variable $f(x^{'}_{j})[a]$. 
To solve for our model's output, we must find the distribution $p(f(x^{'}_{j})[a])$ by solving the intractable integral in equation \ref{eq:intractable_integral}. 
We approach a solution with the same moment matching procedure presented in section \ref{section_prediction}, using the split kernel (equation \ref{eq:split_rbf_kernel}) and probability distribution (equation \ref{eq:split_probability_distribution}) derived above. 
The values of the known pixels have been definitively observed with the assumption of no noise. 
The joint probability distribution of all known pixels can be substituted with a delta function at the observed values: $\delta(x - x^{'}_{j, Kn, \mu})$. 
This allows us to integrate out certain contributions from the known components during the moment matching steps. 
Solving for the mean of $f(x^{'}_{j})[a]$ using these hybrid inputs we get: 
\begin{equation}
    \begin{gathered}
        \mu(x^{'}_{j}) = d_{a, hybrid}^{T} \beta_a \\
        d_{a, hybrid}[i] = k_{a, Kn}(x_{i, Kn}, x^{'}_{j, Kn, \mu}) \cdot  \\
        \frac{\alpha_{a, Rn}^{2}}{\sqrt{|\Sigma_{x^{'}_{j, \sigma}} \Lambda_{a, Rn}^{-1} + I|}} \cdot 
        e^{-\frac{1}{2} v_{i, Rn}^{T} \left( \Sigma_{x^{'}_{j, \sigma}} + \Lambda_{a, Rn} \right)^{-1} v_{i, Rn}}
    \end{gathered}
    \label{eq:hybrid_mean_prediction}
\end{equation}
We use the $\cdot$ operator to denote element wise multiplication. 
$\beta_{a}$ is defined in equation \ref{eq:mean_propagation_random} and 
$v_{i, Rn} = x^{'}_{j, Rn, \mu} - x_{i}$. 
Solving for the variance of $f(x^{'}_{j})[a]$ from the hybrid inputs we get: 
\begin{equation}
    \begin{gathered}
        \Sigma(x^{'}_{j})[a,a] = \alpha_{a}^2 - trace((k_{a}(X, X) + \sigma_{n, a}^2 I)^{-1} Q_{aa, hybrid}) + \\
        \beta_{a}^{T} Q_{aa, hybrid} \beta_{a} - \mu(x^{'}_{j})[a] \\
        Q_{aa, hybrid} = Q_{aa, Kn} \cdot Q_{aa, Rn} \\
        Q_{aa, Kn} = k_{a, Kn}(x^{'}_{j, Kn, \mu}, X_{Kn})^{T} \\
        k_{a, Kn}(x^{'}_{j, Kn, \mu}, X_{Kn}) \in \mathbb{R}^{n \times n} \\
        Q_{aa, Rn}[i, k] = \frac{1}{\sqrt{|R_{Rn}|}} \cdot \\
        k_{a, Rn}(x_{i, Rn}, x^{'}_{j, \mu}) k_{a, Rn}(x_{k, Rn}, x^{'}_{j, \mu}) \cdot e^{\frac{1}{2} z_{ik, Rn}^{T} R_{Rn}^{-1} \Sigma_{x^{'}_{j, \sigma}} z_{ik, Rn}}
    \end{gathered}
    \label{eq:hybrid_var_prediction}
\end{equation}
$R_{Rn} = \Sigma_{x^{'}_{j, \sigma,}}(2\Lambda_{a,Rn}^{-1}) + I$ and $z_{ik, Rn} = \Lambda_{a,Rn}^{-1}v_{i,Rn} + \Lambda_{a, w}^{-1}v_{k,Rn}$.  
We use these equations to predict every pixel in the future image. 
Specific details for each step of the rollout are discussed in section \ref{section_hybrid_input_derivation}
A walkthrough derivation of these equations along with the specifics for each step of the rollout can also be found in section \ref{section_hybrid_input_derivation}.

\subsection{Prediction with all random inputs: Derivations} \label{section_random_input_derivations}
In this section we discuss the derivations for the equations \ref{eq:mean_propagation_random} and \ref{eq:var_propagation_random}. 
These equations predict the mean and variance of the output distribution of a single pixel in a future image $p(f(x^{'}_{j})[a])$.
These predictions are done from input $x^{'}_{j}$ which is a multivariate gaussian random variable $x^{'}_{j} \sim \mathcal{N}(x^{'}_{j, \mu}, \Sigma_{x^{'}_{j, \sigma}})$ defined in section \ref{section_prediction}. 

\paragraph{Mean Prediction Derivation: }
We begin by walking through the derivation of the mean $\mu(x^{'}_{j})[a]$ of the output distribution $p(f(x^{'}_{j})[a])$ presented in equation \ref{eq:mean_propagation_random}. 
For the following derivations we simplify the notation from $p(f(x^{'}_{j})[a]| x^{'}_{j}, X, f(X)[:, a])$ to $p(f(x^{'}_{j})[a]| x^{'}_{j})$ as $X$ and $f(X)[:, a]$ are known quantities. 
We begin our moment matching based derivation by taking the mean of the intractable integral presented in equation \ref{eq:intractable_integral}. 
\begin{equation}
    \begin{gathered}
        \mu(x^{'}_{j})[a] =  E_f[\int_{-\infty}^{\infty} p(f(x^{'}_{j})[a]|x^{'}_{i}) p(x^{'}_{j}) dx^{'}_{i}] \\
        = E_{f, x^{'}_{j}}[p(f(x^{'}_{j})[a]|x^{'}_{j})] \\
        = E_{x^{'}_{j}}[E_f[p(f(x^{'}_{j})[a]|x^{'}_{j})]]
    \end{gathered}
    \label{eq:mean_propagation_random_derivation_1}
\end{equation}
$E_f[p(f(x^{'}_{j})[a]|x^{'}_{j})]$ is the analytical form of the mean during Gaussian Process Regression from equation \ref{eq:GP_mean_pred}. 
Substituting this formula into the above equations we get: 
\begin{equation}
    \begin{gathered}
        \mu(x^{'}_j)[a] = E_{x^{'}_j}[k_a(x^{'}_j, X)[k_a(X, X) + \sigma_{n,a}^2I]^{-1}f(X)[:, a]] \\
    \end{gathered}
    \label{eq:mean_propagation_random_derivation_2}
\end{equation}
We denote $\beta_a \in \mathbb{R}^n$ to be $[k_a(X, X) + \sigma_n^2I]^{-1}f(X)[:, a]]$. 
We denote $d_a \in \mathbb{R}^n$ to be  $E_{x^{’}_{j}}[k_a(x^{'}_{j}, X)]$.
\begin{equation}
    \begin{gathered}
        d_{a}[i] = \int_{-\infty}^{\infty} k_a(x^{'}_{j}, x_{i}) p(x^{'}_{j}) d x^{'}_{j} 
    \end{gathered}
    \label{eq:da_integral}
\end{equation}
\begin{equation}
    \mu(x^{'}_{j})[a] = d_a^T \beta_a = \beta_a^T d_a \in \mathbb{R}
    \label{eq:derivation_mean_prediction}
\end{equation}
Expanding $k_a$ to the RBF kernel equations and $p(x^{'}_{j})$ to the multivariate gaussian pdf, we solve for $d_a$ using \cite{convolution_integral_normal_distrib_rsrc}. 
This gives us the mean prediction equations listed in \ref{eq:mean_propagation_random} and relisted below in equation \ref{eq:appendix_mean_propagation_random}: 

\begin{equation}
    \begin{gathered}
        \mu(x^{'}_{j})[a] = d_a^T \beta_a  \\
        d_a[i] = \frac{\alpha_a^2}{\sqrt{|\Sigma_{x^{'}_{j, \sigma}} \Lambda_a^{-1} + I|}} e^{-\frac{1}{2} v_{i}^{T}(\Sigma_{x^{'}_{j, \sigma}} + \Lambda_a)^{-1}v_{i}} \\
        \beta_a = [k_a(X, X) + \sigma^{2}_{n} I]^{-1} f(X)[:, a]
    \end{gathered}
    \label{eq:appendix_mean_propagation_random}
\end{equation}

\paragraph{Variance Prediction Derivation: }
We now walk through the derivation of the predicted variance $\Sigma({x^{'}_{i}})[a, a] \in \mathbb{R}$. 
Let $\Sigma({x^{'}_{i}}) \in \mathbb{R}^{(p-b)^2 \times (p-b)^2}$ be the covariance matrix of the predicted output, where $\Sigma({x^{'}_{i}})[a, a] \in \mathbb{R}$ is the variance of output $f(x^{'}_{i})[a]$. 
Due to our independence assumptions between outputted pixel distributions, we assert that the covariance between outputs $f(x^{'}_{j})[a]$ and $f(x^{'}_{j})[b]$, representing different output dimensions, $\Sigma({x^{'}_{i}})[a, b] = 0, \; \forall a \neq b$. 
\begin{equation}
    \begin{gathered}
        \Sigma(x^{'}_{j}) = E_{x^{'}_j}\left[ \left( f(x^{'}_{j}) - \mu(x^{'}_{j}) \right)^T \left( f(x^{'}_{j}) - \mu(x^{'}_{j}) \right)  \right]
    \end{gathered}
\end{equation}
This is simplified using the law of total variance. 
\begin{equation}
    \begin{gathered}
        \Sigma(x^{'}_{j})[a,a] = E_{x^{'}_{j}}\left[ var_{f}(f(x^{'}_{j})[a]|x^{'}_{j})\right] + \\
        E_{f, x^{'}_{j}} \left[ f(x^{'}_{j})[a] f(x^{'}_{j})[a]\right] - \mu(x^{'}_{j})[a]^2
    \end{gathered}
    \label{eq:var_propagation_random_presubstitution}
\end{equation}
\begin{equation}
    \begin{gathered}
        E_{f, x^{'}_{j}} \left[ f(x^{'}_{j})[a]f(x^{'}_{j})[a] \right] = \\
        \int_{-\infty}^{\infty} E_{f} \left[ f(x^{'}_{j})[a]|x^{'}_{j}\right] 
        E_{f} \left[ f(x^{'}_{j})[a]|x^{'}_{j}\right] p(x^{'}_{j}) dx^{'}_{j}
    \end{gathered}
\end{equation}
$E_f\left[ f(x^{'}_{j})|x^{'}_{j}\right]$ is the mean output of standard Gaussian Process Regression from equation \ref{eq:GP_mean_pred}. 

Substituting this into the above equations we have: 
\begin{equation}
    \begin{gathered}
        E_{f, x^{'}_{j}}\left[ f(x^{'}_{j})[a]f(x^{'}_{j})[a]\right] \\
        = \int^{\infty}_{-\infty} \beta_a^T k_a(x^{'}_{j}, X)^T k_a(x^{'}_{j}, X) \beta_a p(x^{'}_{j}) dx^{'}_{j} \\
        = \beta_a^T \int^{\infty}_{-\infty} k_a(x^{'}_{j}, X)^T k_a(x^{'}_{j}, X)  p(x^{'}_{j}) dx^{'}_{j} \beta_a
    \end{gathered}
\end{equation}
We define $Q_{aa} =  \int^{\infty}_{-\infty} k_a(x^{'}_{j}, X)^T k_a(x^{'}_{j}, X)  p(x^{'}_{j}) dx^{'}_{j} \in \mathbb{R}^{(p-b)^2 \times (p-b)^2}$. 
$\beta_{a}$ is defined in the above sections. 
This gives us:
\begin{equation}
    \begin{gathered}
         E_{f, x^{'}_{i}} \left[ f(x^{'}_{i})[a]f(x^{'}_{i})[a] \right] = \beta_a^T Q_{aa} \beta_a \\
         Q_{aa}[i,k] = \frac{k_a(x_{i},  x^{'}_{j})k_a({x_{k},  x^{'}_{j}})}{\sqrt{|R|}} e^{z_{ik}^T \mathbb{R}^{-1} \Sigma_{x^{'}_{j, \sigma}} z_{ik}}
    \end{gathered}
    \label{eq:second_term_var_prop_random}
\end{equation}
$z_{ik} = \Lambda_a^{-1}v_i + \Lambda_a^{-1}v_k$ and $R = \Sigma_{x^{'}_{j, \sigma}}[\Lambda_a^{-1} + \Lambda_a^{-1}] + I$ where  $I \in \mathbb{R}^{n \times n}$ is the identity matrix. 

In the first term of equation \ref{eq:var_propagation_random_presubstitution}, $ E_{x^{'}_{j}} \left[ var_{f}(f(x^{'}_{j})[a]|x^{'}_{j}) \right]$, $var_{f}(f(x^{'}_{j})[a]|x^{'}_{j})$ is the variance output of Gaussian Process Regression from equation \ref{eq:GP_var_pred}.
Simplifying this term we get: 
\begin{equation}
    \begin{gathered}
        E_{x^{'}_{j}} \left[ var_{f}(f(x^{'}_{j})[a]|x^{'}_{j}) \right] = \alpha^2 - \\
        trace \left( (k_a(X, X) + \sigma_{n,a}^{2}I)^{-1}Q_{aa}\right)
    \end{gathered}
    \label{eq:first_term_var_prop_random}
\end{equation}
We substitute the equations from \ref{eq:first_term_var_prop_random}, \ref{eq:second_term_var_prop_random} and \ref{eq:mean_propagation_random} into equation \ref{eq:var_propagation_random_presubstitution} to compute the variance for each outputted pixel corresponding to output dimensions $a \in [0, (p-b)^2-1]$. 
This results in the variance prediction formula presented in equations \ref{eq:var_propagation_random} and \ref{eq:var_propagation_random_Qmatrix} and relisted below as equations \ref{eq:appendix_var_propagation_random} and \ref{eq:appendix_var_propagation_random_Qmatrix}

\begin{equation}
    \begin{gathered}
        \Sigma(x^{'}_{j})[a,a] = \alpha_{a}^2 - trace ((k_a(X, X) + \\
        \sigma_{n, a}^{2} I)^{-1} Q_{aa} )) + 
        \beta_a^T Q_{aa} \beta_a - \mu(x^{'}_{j})[a]
    \end{gathered}
    \label{eq:appendix_var_propagation_random}
\end{equation}
\begin{equation}
    \begin{gathered}
        Q_{aa}[i,k] = \frac{k_a(x_{i}, x^{'}_{j}) k_a(x_{k}, x^{'}_{j})}{\sqrt{|R|}} e^{\frac{1}{2}z_{ik}^T R^{-1}\Sigma_{x^{'}_{j, \sigma}} z_{ik}}
    \end{gathered}
    \label{eq:appendix_var_propagation_random_Qmatrix}
\end{equation}

\subsection{Prediction with Hybrid and Fully Known inputs: Derivations and Additional Details}
\label{section_hybrid_input_derivation}
In this section we discuss the derivations for the equations \ref{eq:hybrid_mean_prediction} and \ref{eq:hybrid_var_prediction}. 
These equations predict the mean and variance of the output distribution of a single pixel in a future image $p(f(x^{'}_{j})[a])$.
These predictions are done from input $x^{'}_{j}$ which is composed of two random variables $x^{'}_{j, Rn}$ and $x^{'}_{j,Kn}$ explained in section \ref{section_hybrid_input_derivation}.

\paragraph{Mean Prediction Derivation: }
In this section we discuss the derivation of the mean prediction $\mu(x^{'}_{j})[a]$ of the output distribution $p(f(x^{'}_{j})[a])$ when using a combination of known and random input images, for the first $3$ predictions of a rollout. 
We show the derivation for equation \ref{eq:hybrid_mean_prediction}. 
Since this is a special case of prediction from all random inputs we begin our derivation from the derivation of the mean prediction equations for all random inputs in section \ref{section_random_input_derivations}
We begin our derivation from equations
\ref{eq:mean_propagation_random_derivation_2}, \ref{eq:da_integral} and \ref{eq:derivation_mean_prediction}. 

In this derivation we use the split kernel and probability density functions in equations \ref{eq:split_rbf_kernel} and \ref{eq:split_probability_distribution} to deal with the hybrid, random and known, nature of the inputs.
The joint probability distribution of all known pixels can be substituted with a delta function at the known values: $\delta(x - x^{'}_{j, Kn, \mu})$. 
This allows us to integrate out certain contributions from the known components. 
$\beta_a$, being a constant, remains unchanged and we re derive $d_a$ as $d_{a, hybrid}$. 
\begin{equation}
    \begin{gathered}
        d_{a,hybrid}[i] =  E_{x^{'}_j}[k_a(x^{'}_j, X[i])] \\
        = E_{x^{'}_{j, Rn}, x^{'}_{j, Kn}}[k_{a, Rn}(x^{'}_{j, Rn}, X_{Rn}[i]) k_{a, Kn}(x^{'}_{j, Kn}, X_{Kn}[i]) ] \\
        = E_{x^{'}_{j, Rn}}[k_{a, Rn}(x^{'}_{j, Rn}, X_{Rn})] E_{x^{'}_{j, Kn}}[k_{a, Kn}(x^{'}_{j, Kn}, X_{Kn})] \\
        = \int_{-\infty}^{\infty} k_{a, Kn}(x^{'}_{j, Kn}, X_{Kn}) \delta(x^{'}_{j, Kn, \mu} - x^{'}_{j,Kn}) dx^{'}_{j, Kn} \\
        \int_{-\infty}^{\infty} k_{a, Rn}(x^{'}_{j, Rn}, X_{Rn}) p(x^{'}_{j, Rn}) dx^{'}_{Rn}
    \end{gathered}
\end{equation}
Solving this yields the mean prediction for the third rollout given in equation \ref{eq:hybrid_mean_prediction} relisted below as equation \ref{eq:appendix_hybrid_mean_prediction}: 
\begin{equation}
    \begin{gathered}
        \mu(x^{'}_{j}) = d_{a, hybrid}^{T} \beta_a \\
        d_{a, hybrid}[i] = k_{a, Kn}(x_{i, Kn}, x^{'}_{j, Kn, \mu}) \cdot  
        \frac{\alpha_{a, Rn}^{2}}{\sqrt{|\Sigma_{x^{'}_{j, \sigma}} \Lambda_{a, Rn}^{-1} + I|}} \cdot \\
        e^{-\frac{1}{2} v_{i, Rn}^{T} \left( \Sigma_{x^{'}_{j, \sigma}} + \Lambda_{a, Rn} \right)^{-1} v_{i, Rn}}
    \end{gathered}
    \label{eq:appendix_hybrid_mean_prediction}
\end{equation}

\paragraph{Variance Prediction Derivation: }
Here we discuss the derivation of the variance $\Sigma({x^{'}_{i}})[a, a] \in R$ in equation \ref{eq:hybrid_var_prediction} from hybrid and random inputs. 
We follow the method outlined in the derivation for all random inputs. 
We use the split kernel and probability density functions in equations \ref{eq:split_rbf_kernel} and \ref{eq:split_probability_distribution} to separately deal with the random and known components of the inputs.
With this we arrive at an identical formulation to the case with all random inputs where $Q_{aa, hybrid}$ is used in place of $Q_{aa}$. 
\begin{equation}
    \begin{gathered}
        Q_{aa,hybrid} = E_{x^{'}_{j}} [k_{a}(x^{'}_{j}, X)^T k_{a}(x^{'}_{j}, X)] \\
        = E_{x^{'}_{j, Rn}}[k_{a, Rn}(x^{'}_{j, Rn}, X_{Rn})^{T}k_{a, Rn}(x^{'}_{j, Rn}, X_{Rn})] \cdot \\
        E_{x^{'}_{j, Kn}}[k_{a, Kn}(x^{'}_{j, Kn}, X_{Kn})^{T} k_{a, Kn}(x^{'}_{j, Kn}, X_{Kn})] \\
        = \int_{-\infty}^{\infty} k_{a, Kn}(x^{'}_{j, Kn}, X_{Kn})^{T}k_{a, Kn}(x^{'}_{j, Kn}, X_{Kn}) \\
        \delta(x^{'}_{j, Kn} - x^{'}_{j,Kn, \mu}) dx^{'}_{j, Kn} \\
        \int_{-\infty}^{\infty} k_{a, Rn}(x^{'}_{j, Rn}, X_{Rn})^{T} k_{a, Rn}(x^{'}_{j, Rn}, X_{Rn}) p(x^{'}_{j, Rn}) dx^{'}_{Rn} 
    \end{gathered}
\end{equation}
Here $\cdot$ denotes the element wise multiplication operator. 
The integrals with the multivariate gaussian pdfs result in the same solution as elaborated in the random variance derivation. 
Solving this yields the variance prediction given in equation \ref{eq:hybrid_var_prediction} relisted below as equation \ref{eq:appendix_hybrid_var_prediction}:
\begin{equation}
    \begin{gathered}
        \Sigma(x^{'}_{j})[a,a] = \alpha_{a}^2 - \\
        trace((k_{a}(X, X) + \sigma_{n, a}^2 I)^{-1} Q_{aa, hybrid}) + \\
        \beta_{a}^{T} Q_{aa, hybrid} \beta_{a} - \mu(x^{'}_{j})[a] \\
        Q_{aa, hybrid} = Q_{aa, Kn} \cdot Q_{aa, Rn} \\
        Q_{aa, Kn} = k_{a, Kn}(x^{'}_{j, Kn, \mu}, X_{Kn})^{T} k_{a, Kn}(x^{'}_{j, Kn, \mu}, X_{Kn}) \in \mathbb{R}^{n \times n} \\
        Q_{aa, Rn}[i, k] = \frac{1}{\sqrt{|R_{Rn}|}} \cdot k_{a, Rn}(x_{i, Rn}, x^{'}_{j, \mu}) k_{a, Rn}(x_{k, Rn}, x^{'}_{j, \mu}) \cdot \\
        e^{\frac{1}{2} z_{ik, Rn}^{T} R_{Rn}^{-1} \Sigma_{x^{'}_{j, \sigma}} z_{ik, Rn}}
    \end{gathered}
    \label{eq:appendix_hybrid_var_prediction}
\end{equation}

\paragraph{Rollout Discussion: }
In this section we discuss the composition of our inputs and additional details of each step in our predictive rollout. 
In a predictive rollout we are predicting $T$ time steps into future starting from $3$ known, consecutive input images $[z_{i}, z_{i+1}, z_{i+2}]$. 
With each prediction we predict a single time step into the future before incorporating our prediction into our next set of inputs. 
We continue this process until we predict the desired number of time steps. 

\textit{First Step: }
The first step of the predictive rollout predicts $z_{i+3}$ from input images $[z_{i}, z_{i+1}, z_{i+2}]$. 
For this first prediction all the input images are known quantities. 
As a result for each test input $x^{'}_{j} \: j \in [0, m-1]$, the entire test input is known, $x^{'}_{j} = x^{'}_{j, Kn}$, and $x^{'}_{j, Rn}$ does not exist. 
$ x^{'}_{j, Kn, \mu} \in \mathbb{R}^{3p^2}$ is formed in a manner identical to the  training inputs. 
When plugging these inputs into the hybrid mean prediction equation \ref{eq:hybrid_mean_prediction} and variance prediction equation \ref{eq:hybrid_var_prediction} we remove the random components of the equations giving us: 
\begin{equation}
    d_{a, hybrid}[i] = k_{a, Kn}(x_{i, Kn}, x^{'}_{j, Kn, \mu})
\end{equation}
\begin{equation}
    Q_{aa, hybrid} = Q_{aa, Kn}
\end{equation}  
Substituting these back into the equations \ref{eq:hybrid_mean_prediction} and \ref{eq:hybrid_var_prediction} we get the formulas to predict a single output dimension of for a single test input for the first prediction. 
These equations equivalent to the basic Gaussian Process Regression equations for mean and variance prediction. 
Using these formulas we predict the distribution for every pixel in $z_{i+3}$ from the $m$ test inputs. 
This gives us the final predicted image $z_{i+3}$ which is stored as a mean, variance image tuple $(M_{i+3}, V_{i+3})$.

\textit{Second Step: }
The second step of the predictive rollout predicts $z_{i+4}$ from input images $[z_{i+1}, z_{i+2}, z_{i+3}]$. 
$z_{i+3}$ is a random variable, from the first prediction, represented by the mean and variance image tuple  $(M_{i+3}, V_{i+3})$.
$z_{i+1}$ and $z_{i+2}$ are known. 
When constructing each test input, 
$x^{'}_{j, Kn, \mu}$ is constructed from flattened and concatenated patches of $[z_{i+1}, z_{i+2}]$. 
$x^{'}_{j, \mu}$ is constructed from flattened patches of $M_{i+3}$ and  $x^{'}_{j, \sigma}$ is constructed from flattened patches of $V_{i+3}$ to create the random input $x^{'}_{j, Rn}$. 
These components together form a single test input $x^{'}_{j}$. 
We plug these inputs into the hybrid mean prediction equation \ref{eq:hybrid_mean_prediction} and variance prediction equation \ref{eq:hybrid_var_prediction} to compute the model's output distribution for a single output dimension. 
We repeat this to predict the output distributions for each pixel in the future image $z_{i+4}$ which is stored as a mean and variance image tuple  $(M_{i+3}, V_{i+3})$.

\textit{Third Step: }
The third step of the predictive rollout predicts $z_{i+5}$ from input images $[z_{i+2}, z_{i+3}, z_{i+4}]$. 
$z_{i+3}$ and $z_{i+4}$ are random variables, from the first and second predictions, represented by the mean and variance image tuples  $(M_{i+3}, V_{i+3}), (M_{i+4}, V_{i+4})$.
$z_{i+2}$ is known. 
When constructing each test input, 
$x^{'}_{j, Kn, \mu}$ is constructed from flattened patches of $[z_{i+2}]$. 
$x^{'}_{j, \mu}$ is constructed from flattened concatenated patches of $M_{i+3}, M_{i+4}$ and  $x^{'}_{j, \sigma}$ is constructed from flattened patches of $V_{i+3}, V_{i+4}$ to create the random input $x^{'}_{j, Rn}$. 
These components together form a single test input $x^{'}_{j}$. 
We plug these inputs into the hybrid mean prediction equation \ref{eq:hybrid_mean_prediction} and variance prediction equation \ref{eq:hybrid_var_prediction} to compute the model's output distribution for a single output dimension. 
We repeat this to predict the output distributions for each pixel in the future image $z_{i+5}$ which is stored as a mean and variance image tuple  $(M_{i+5}, V_{i+5})$. 

\textit{Fourth Step and Onwards: }
The third step of the predictive rollout predicts $z_{i+6}$ from input images $[z_{i+3}, z_{i+4}, z_{i+5}]$.
For this and all subsequent predictions, all the input images are random variables outputted by our model. 
To predict the future image we utilize the approach detailed in the section \ref{section_prediction} on Prediction with all random inputs. 

\subsection{Training Input Creation Graphic}\label{section_appendix_training_data_creation}
\begin{figure*}[h!]
    \centering
    \includegraphics[width=\textwidth]{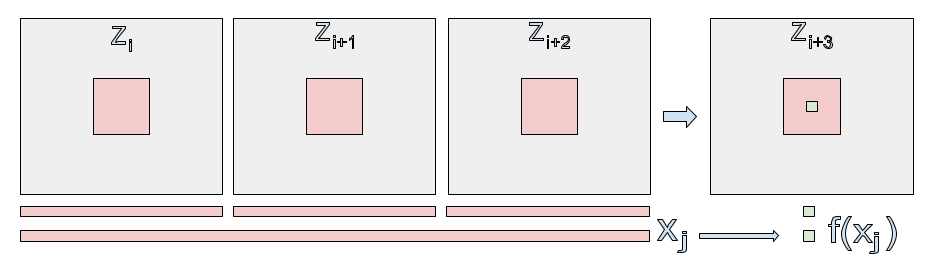}
    \caption{Graphic to visualize the process of generating a training data point from a sequence of $4$ consecutive video frames. 
    Each grey figure represents a labelled video frame. 
    The red squares represent the patches that are sampled to create the input output pair. 
    In the fourth image the green pixel represents the output patch after cropping the patch with patch border $b$. 
    The patches from the first $3$ images are flattened and concatenated to form the input. 
    The patch from the fourth image is flattened and used as the output. 
    In this case the output patch is a single pixel. }
    \label{fig:training_data_creation}
\end{figure*}

The Figure \ref{fig:training_data_creation} graphically demonstrates the process of creating a training datapoint from patches of $4$ consecutive video frames. 

\subsection{Additional Experiments and Details}
\label{section_predictive_comparison_additional_details}

\begin{figure*}[t]
    \centering
    \begin{subfigure}[b]{1.0\textwidth}
        \centering
        \includegraphics[height=1.35cm]{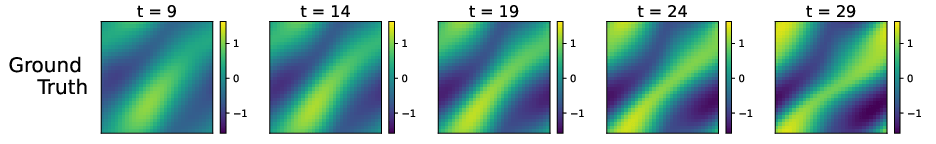}
        \vspace{-1mm}
        \caption{Ground Truth Image Frames}
        \label{fig:ground_truth_sequence_prediction}
    \end{subfigure}
    \vfill
    \vspace{-3mm}
    \begin{subfigure}[b]{0.7\textwidth}
        \centering
        \includegraphics[height=4.25cm]{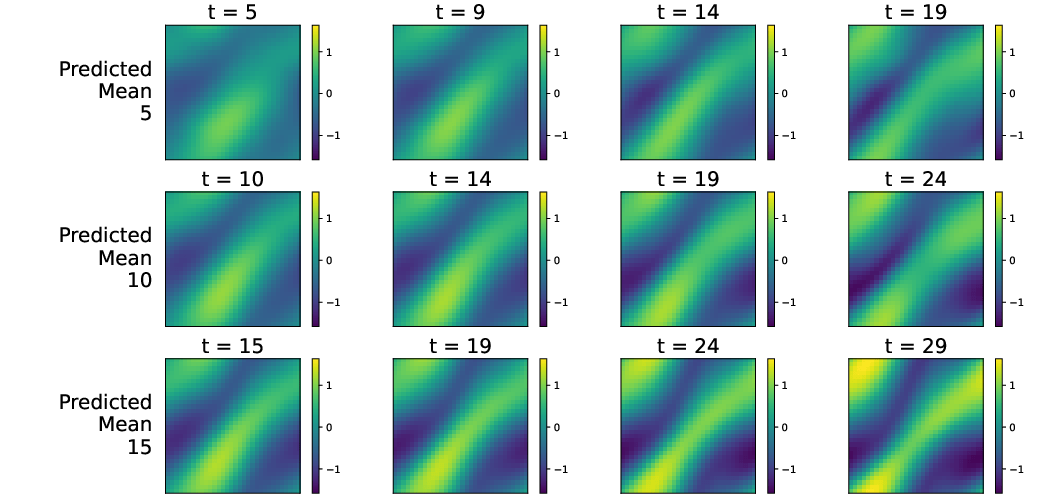}
        \vspace{-1mm}
        \caption{Sequence Prediction Mean Images}
        \label{fig:sequence_prediction_images}
    \end{subfigure}
    \hfill
    \begin{subfigure}[b]{0.29\textwidth}
        \centering
        \includegraphics[width=\textwidth]{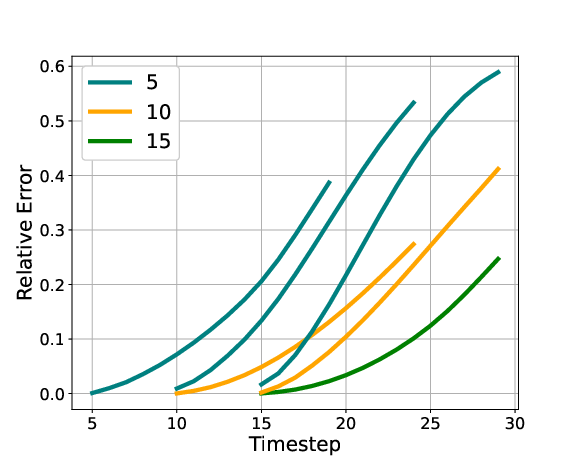}
        \vspace{-1mm}
        \caption{Sequence Prediction Errors}
        \label{fig:sequence_prediction_errors}
    \end{subfigure}
    \vspace{-2mm}
    \caption{ 
    Sequential Prediction Experiment:
    Our model, trained using frames $[z_{0}, \dots, z_{t_{0}-1}]$, is used to predict frames $[z_{t_{0}}, \dots, z_{t_{0}+15}]$ of a 2D Navier Stokes simulation.
    Prediction results for $t=5,10,15$ are shown along the rows of \ref{fig:sequence_prediction_images}, respectively. 
    Figure \ref{fig:ground_truth_sequence_prediction} displays ground truth images. 
    Figure \ref{fig:sequence_prediction_images} shows the predicted mean images. 
    Figure \ref{fig:sequence_prediction_errors} displays a graph of the relative error of the predicted means. 
    In Figure \ref{fig:sequence_prediction_errors} we show additional error results where we start predictive rollouts with the models trained with $5, 10$ images from later time steps. 
    This provides a fair experiment to show the predictive improvement resulting from incorporating recent data into our model. 
    }
    \label{fig:sequence_prediction}
\end{figure*}

\subsubsection{Predictive Comparison Experiment: Additional Details}
In this section we highlight additional details on the methodology used to generate the results for the 'Predictive Comparison' Experiment in Section \ref{section_experiments_and_results}. 
To compare all three methods, we predict frames $[z_{10}, \dots, z_{24}]$ given frames $[z_{0}, \dots, z_{9}]$. 
For our method we train our model using  $[z_{0}, \dots, z_{9}]$ and begin our prediction with input images $[z_{7}, z_{8}, z_{9}]$, identical to the approach outlined in the ‘Forward Prediction’ Experiment. 

The FNO-2d-time model convolves across the two spatial dimensions to predict a single future image with a recurrent structure in time. 
This model uses a rollout method to predict longer video sequences. 
The model predicts one future frame at a time and incorporates its last prediction into its next input. 
We continue this until we have predicted the desired number of frames. 
The model is structured to take an input of three images and predict a single future image. 
We train this method in a manner similar to ours. 
The training data is created using the first $10$ frames $[z_{0}, \dots, z_{9}]$.
These frames are separated into data points of $4$ consecutive images $[z_{i}, \dots, z_{i+3}] ; \forall i \in [0, 6]$, where the first three images form the input and the fourth is the output.
The model is trained over 20 epochs.  
The trained model is then used to predict $15$ frames $[z_{10}, \dots, z_{24}]$ of the same sequence. 

The FNO-3d model is a neural network that convolves in space and time to directly output several frames from a set of input frames.
We train this model using the first $25$ frames from two unique sequences, generated using the same simulation parameters. 
The first $10$ images are used as the inputs, with the remaining $15$ serving as the training outputs. 
Once trained over 20 epochs this model is given a set of $10$ consecutive frames  $[z_{0}, \dots, z_{9}]$ to predict the next $15$: $[z_{10}, \dots, z_{24}]$. 

The results of this comparison are shown in fig. \ref{fig:predictive_comparison_experiment}.

\subsubsection{Average Result Metrics}
\label{section_average_result_metrics}
\begin{figure*}[h!]
    \centering
    \begin{subfigure}[b]{0.49\textwidth}
        \centering
        \includegraphics[width=\textwidth]{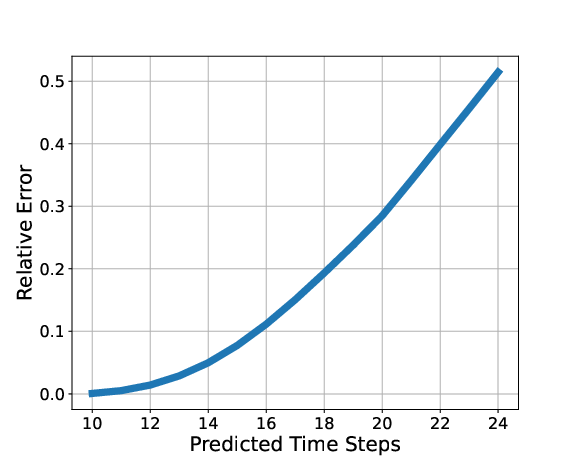}
        \caption{Average Relative Errors}
        \label{fig:avg_predictive_comparison_errors}
    \end{subfigure}
    \hfill
    \begin{subfigure}[b]{0.49\textwidth}
        \centering
        \includegraphics[width=\textwidth]{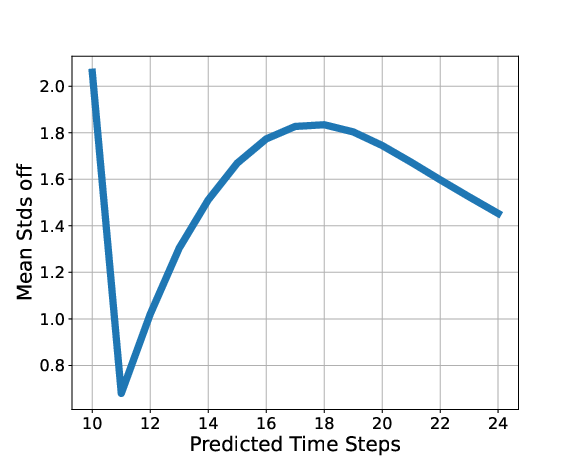}
        \caption{Average Mean Standard Deviations off}
        \label{fig:avg_mean_stds_off}
    \end{subfigure}
    \caption{Averaged Metrics: Figure \ref{fig:avg_predictive_comparison_errors} Shows the average relative error metrics using our method over 100 results on separate video sequences. Figure \ref{fig:avg_mean_stds_off} Shows the average mean standard deviations off the predicted image is from the ground truth using the mean and variance generated using our method. This result is also averaged over predictions on 100 different video sequences. To generate the above results our model was trained on frames $[z_{0}, \dots, z_{9}]$ of each sequence and used to predict the next $15$ frames, $[z_{10}, \dots, z_{24}]$}
    \label{fig:average_result_figure}
\end{figure*}

We showcase the average relative error and average mean standard deviations off of our model's predictions evaluated on $100$ separate video sequences in Figure \ref{fig:average_result_figure}. 
For each video sequence our model is trained using the first $10$ frames $[z_{0}, \dots, z_{9}]$ and used to predict the next $15$ frames, $[z_{10}, \dots, z_{24}]$. 
We use the parameters discussed in the 'Forward Prediction' experiment. 

\subsubsection{Sequential Prediction Experiment: } 
In this experiment we examine the benefits of incorporating recent data into our model.
We incrementally train models with the first $5,10$ and $15$ images of a video sequences. 
Each of these models is used to predict $15$ frames into the future, starting from their last training image.
In Figure \ref{fig:sequence_prediction} we can visually see the improvement in prediction accuracy as we update our models. 
In the relative error graph (Figure  \ref{fig:sequence_prediction_errors}) we also show the results of starting predictive rollouts for the lower data models, trained with $5$ and $10$ images, later in the sequence.
This provides a fair evaluation to compare the impact of adding recent data, by mitigating the added error compounded as a result of the predictive rollouts. 
The graph shows a large improvement in accuracy as we incorporate data into our model. 

\subsubsection{Additional Visual Results}
\begin{figure*}[t]
    \centering
    \begin{subfigure}{0.69\textwidth}
        \centering
        \includegraphics[height=4.8cm]{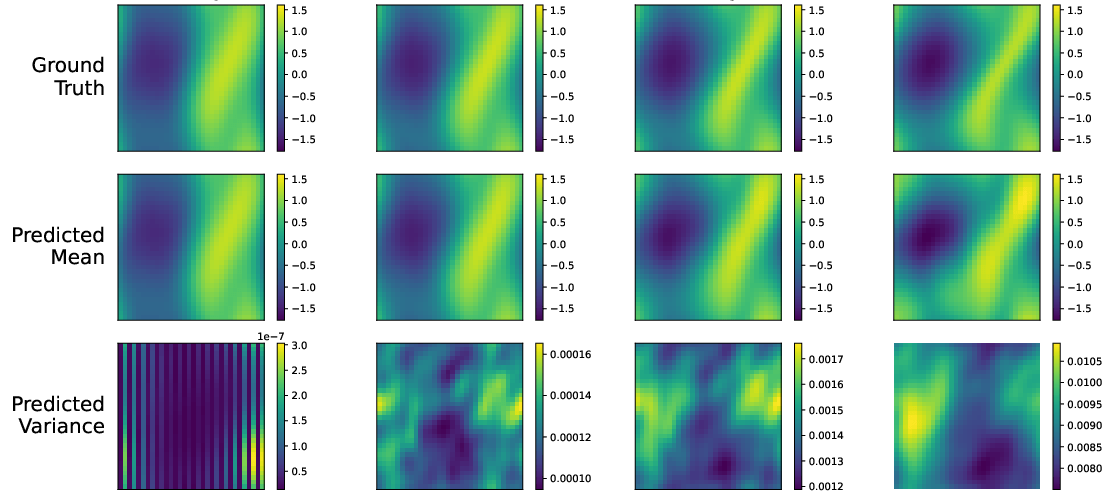}
        \vspace{-1mm}
        \caption{Forward Prediction: Image Results}
        \label{fig:additional_prediction_images}
    \end{subfigure}
    \begin{subfigure}{0.3\textwidth}
        \centering
        \begin{subfigure}{1.0\textwidth}
            \centering
            \includegraphics[height=2.15cm]{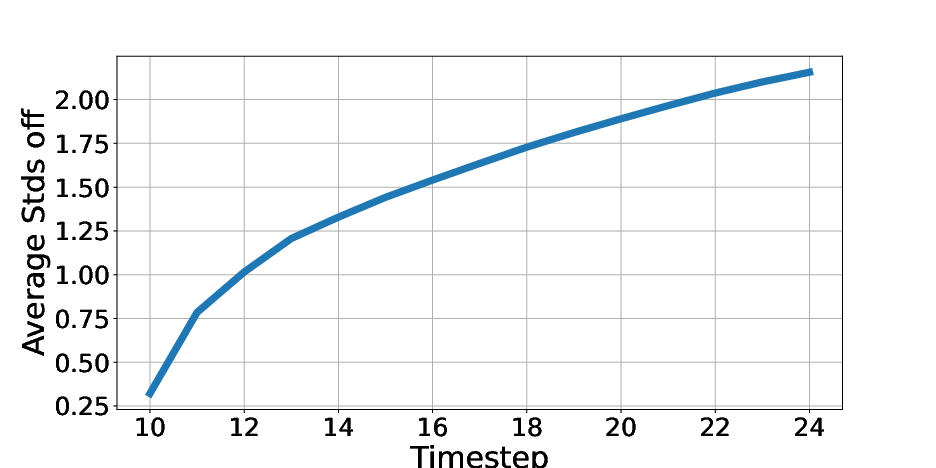}
            \vspace{-1mm}
            \caption{Relative Error Graph}
            \label{fig:additional_prediction_errors}
        \end{subfigure}
        \\
        \vspace{-2mm}
        \begin{subfigure}{1.0\textwidth}
            \centering
            \includegraphics[height=2.15cm]{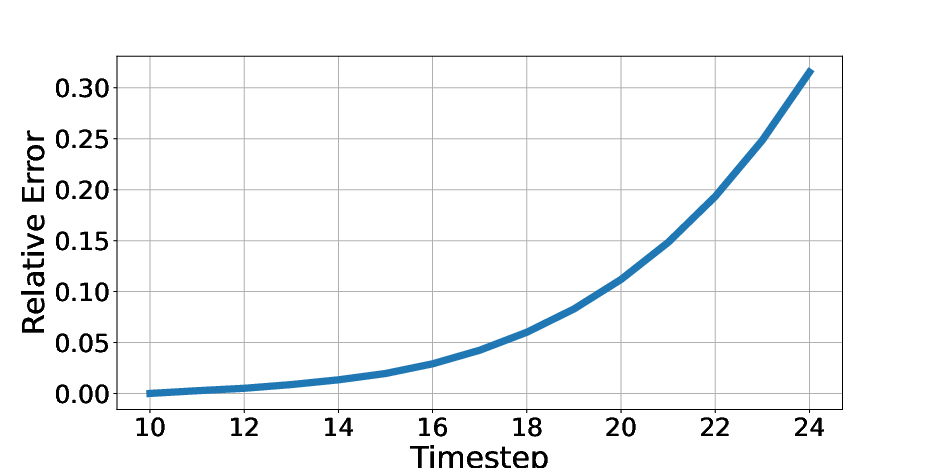}
            \vspace{-1mm}
            \caption{Mean Standard Deviations off}
            \label{fig:additional_prediction_stdsoff}
        \end{subfigure}
    \end{subfigure}
    \vspace{-2mm}
    \caption{Forward Prediction Experiment: Additional Experiment $1$:
    Our model, trained using frames $[z_{0}, \dots, z_{9}]$, is used to predict frames $[z_{10}, \dots, z_{24}]$ of a 2D Navier Stokes simulation. 
    Figure \ref{fig:additional_prediction_images} shows the ground truth, predicted mean and variance images. 
    Figure \ref{fig:additional_prediction_errors} and Figure \ref{fig:additional_prediction_stdsoff} show graphs of the prediction's relative error and mean standard deviations off. 
}
    \label{fig:additional_prediction_experiment}
\end{figure*}

\begin{figure*}[t]
    \centering
    \begin{subfigure}{0.69\textwidth}
        \centering
        \includegraphics[height=4.8cm]{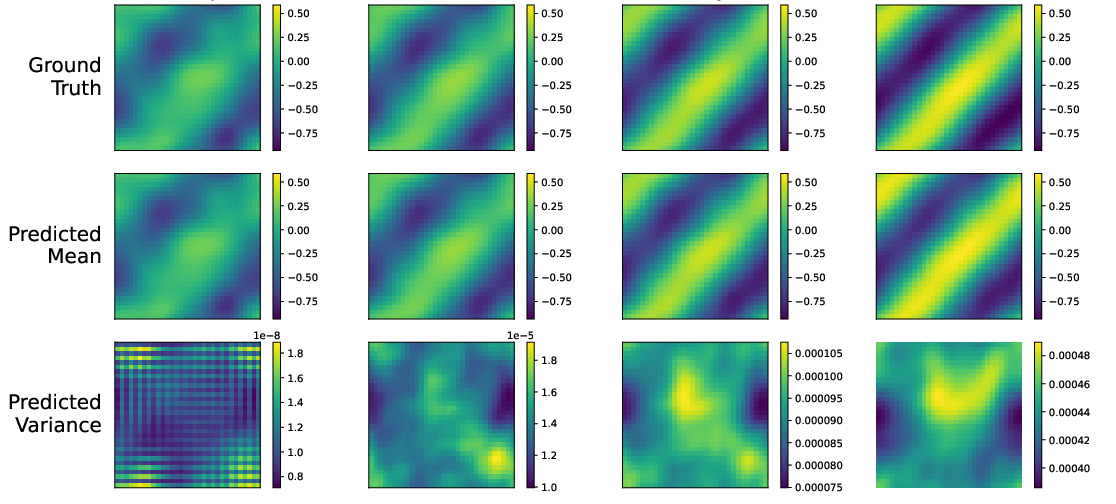}
        \vspace{-1mm}
        \caption{Forward Prediction: Image Results}
        \label{fig:additional_prediction_images2}
    \end{subfigure}
    \begin{subfigure}{0.3\textwidth}
        \centering
        \begin{subfigure}{1.0\textwidth}
            \centering
            \includegraphics[height=2.15cm]{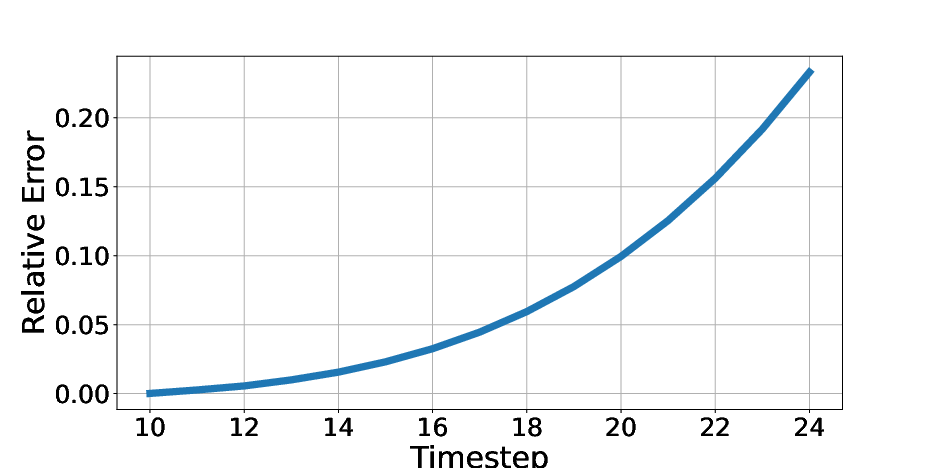}
            \vspace{-1mm}
            \caption{Relative Error Graph}
            \label{fig:additional_prediction_errors2}
        \end{subfigure}
        \\
        \vspace{-2mm}
        \begin{subfigure}{1.0\textwidth}
            \centering
            \includegraphics[height=2.15cm]{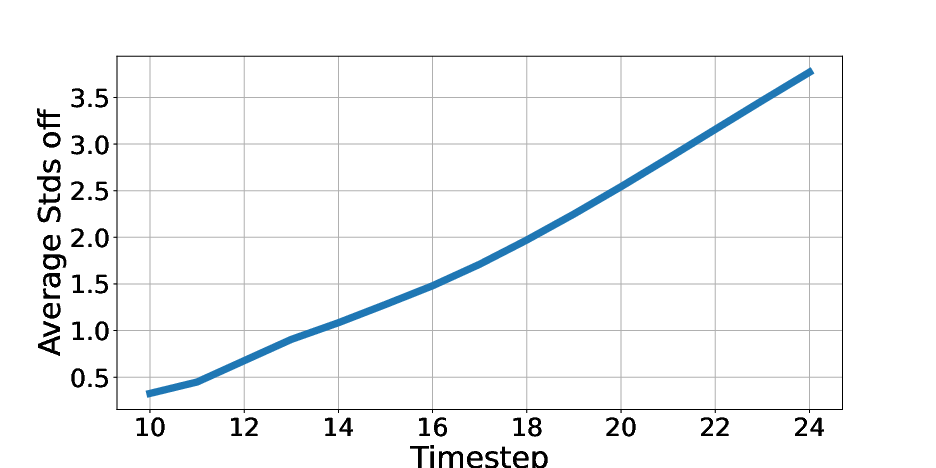}
            \vspace{-1mm}
            \caption{Mean Standard Deviations off}
            \label{fig:additional_prediction_stdsoff2}
        \end{subfigure}
    \end{subfigure}
    \vspace{-2mm}
    \caption{Forward Prediction Experiment: Additional Experiment $2$:
    Our model, trained using frames $[z_{0}, \dots, z_{9}]$, is used to predict frames $[z_{10}, \dots, z_{24}]$ of a 2D Navier Stokes simulation. 
    Figure \ref{fig:additional_prediction_images2} shows the ground truth, predicted mean and variance images. 
    Figure \ref{fig:additional_prediction_errors2} and Figure \ref{fig:additional_prediction_stdsoff2} show graphs of the prediction's relative error and mean standard deviations off. 
}
    \label{fig:additional_prediction_experiment2}
\end{figure*}

Figures \ref{fig:additional_prediction_experiment} and \ref{fig:additional_prediction_experiment2} show additional results for predictions on different Navier Stokes simulations using the "Forward Prediction Experiment" detailed in section \ref{section_experiments_and_results}. 

\end{document}